\documentclass[letterpaper, 10 pt, conference]{ieeeconf}  %

\IEEEoverridecommandlockouts                              %

\usepackage[table]{xcolor}
\usepackage{hhline}
\usepackage{xspace}
\usepackage{algorithm}
\usepackage{booktabs}
\usepackage{graphicx}
\usepackage{algpseudocode}
\usepackage{siunitx}
\usepackage{svg}
\usepackage{float}
\usepackage{subcaption}
\usepackage{inconsolata}
\usepackage{xcolor}
\usepackage{bbding}
\usepackage{amsfonts}
\usepackage{wrapfig}
\usepackage{amsmath,bm}
\usepackage{hyperref}
\usepackage{multirow}
\usepackage{xcolor}
\usepackage{amsmath}
\usepackage{amssymb}
\usepackage{multirow}
\usepackage{booktabs}
\usepackage{graphicx}
\usepackage{mdframed}
\usepackage{caption}
\usepackage{verbatim}
\usepackage{xspace}
\usepackage{paralist}
\usepackage{cuted}
\usepackage{hyperref}

\usepackage[noadjust]{cite}

\usepackage{listings}
\usepackage{xcolor}  %

\usepackage{listings}
\newcommand{\algname}{Unsupervised Affordance Distillation\xspace}
\newcommand{\algabrvname}{UAD\xspace}

\title{\LARGE \bf
UAD: Unsupervised Affordance Distillation\\for Generalization in Robotic Manipulation
}

\author{Yihe Tang, Wenlong Huang, Yingke Wang, Chengshu Li, Roy Yuan, \\ Ruohan Zhang, Jiajun Wu, Li Fei-Fei\\
Stanford University %
}

\begin{document}

\twocolumn[{%
\renewcommand\twocolumn[1][]{#1}%
\maketitle
\vspace{-3em}
\begin{center}
    \centering
    \captionsetup{type=figure}
    \includegraphics[width=1.0\textwidth]{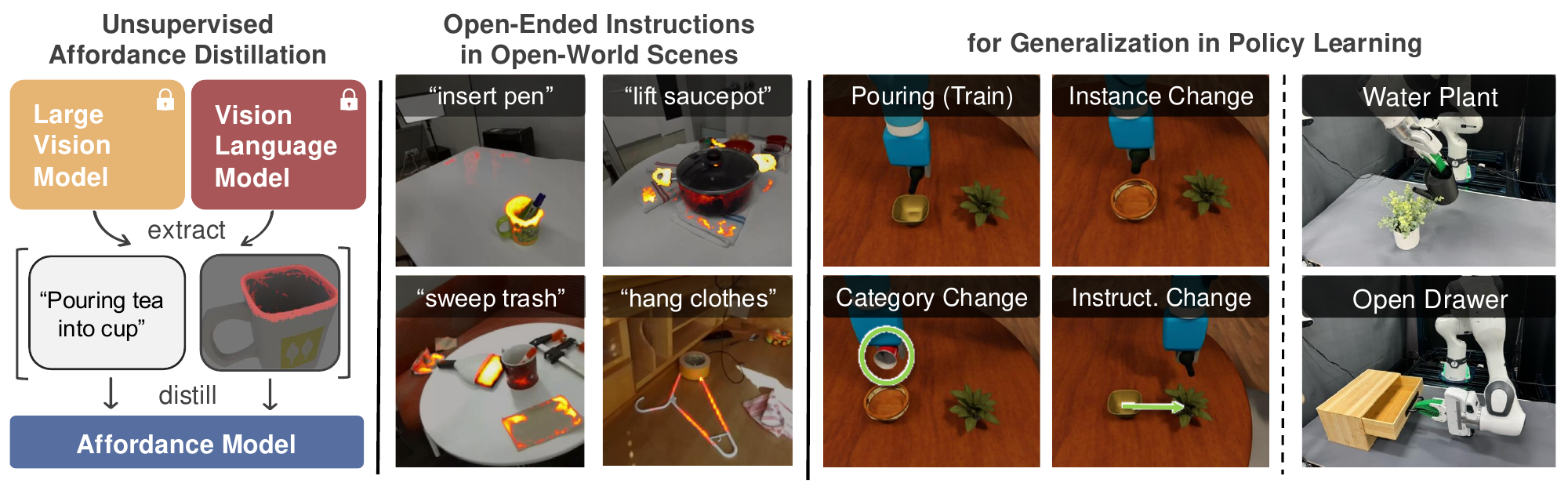}
    \captionof{figure}{
    \textbf{\algname (\algabrvname)} extracts affordance annotations from large pre-trained models and distills them into a task-conditioned affordance model, which is capable of predicting fine-grained affordance in open-world scenes with open-ended instructions, enabling diverse generalization properties in downstream policy learning.}
    \label{fig:teaser}
\end{center}%
}]

\thispagestyle{empty}
\pagestyle{empty}

\begin{abstract}

Understanding fine-grained object affordances is imperative for robots to manipulate objects in unstructured environments given open-ended task instructions.
However, existing methods of visual affordance predictions often rely on manually annotated data or conditions only on a predefined set of tasks.
We introduce \algname~(\algabrvname), a method for distilling affordance knowledge from foundation models into a task-conditioned affordance model \emph{without any manual annotations}.
By leveraging the complementary strengths of large vision models and vision-language models, \algabrvname automatically annotates a large-scale dataset with detailed $<$instruction, visual affordance$>$ pairs.
Training only a lightweight task-conditioned decoder atop frozen features, \algabrvname exhibits notable generalization to in-the-wild robotic scenes and to various human activities, despite only being trained on rendered objects in simulation.
Using affordance provided by \algabrvname as the observation space, we show an imitation learning policy that demonstrates promising generalization to unseen object instances, object categories, and even variations in task instructions after training on as few as 10 demonstrations.
Project website: \href{https://unsup-affordance.github.io/}{unsup-affordance.github.io/}.

\end{abstract}

\begin{figure*}[t]
    \centering
    \includegraphics[width=0.85\linewidth]{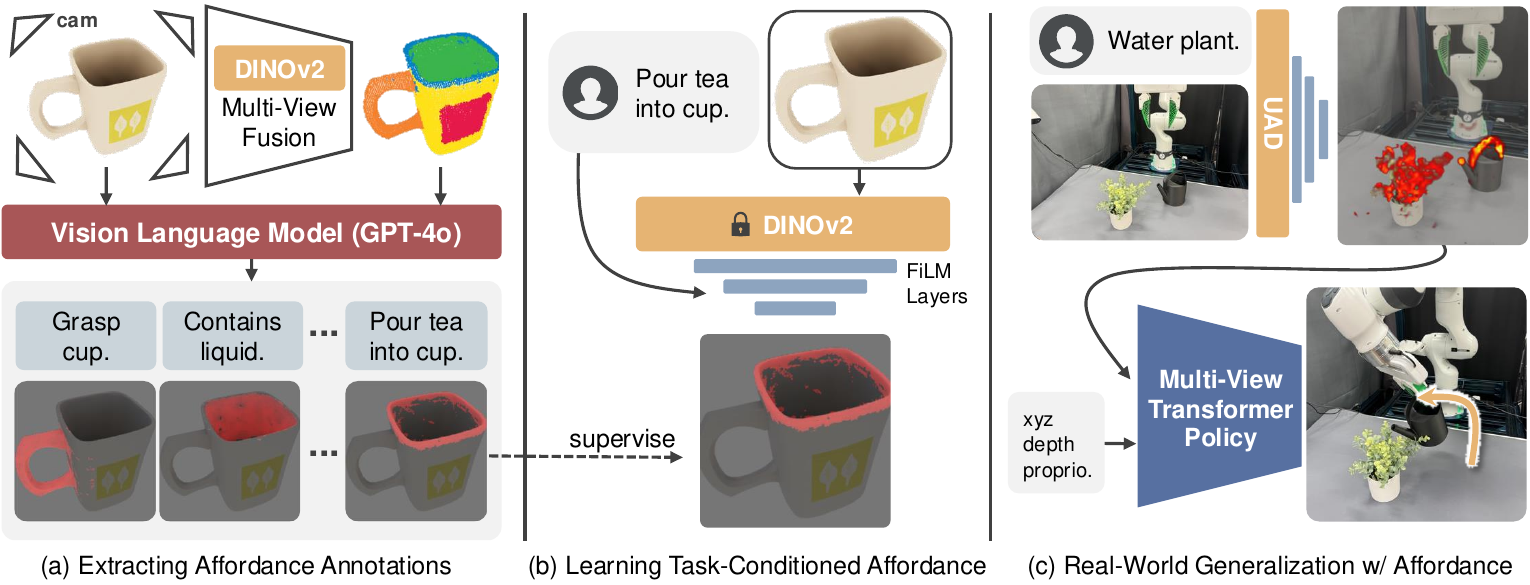}
    \caption{
    \textbf{Overview of Unsupervised Affordance Distillation (UAD).}
    Using renderings of 3D objects, we first perform multi-view fusion of DINOv2 features and clustering to obtain fine-grained semantic regions of objects, which are then fed to VLM for proposing relevant tasks and corresponding regions \textbf{(a)}. The extracted affordance is then distilled by training a language-conditioning FiLM atop frozen DINOv2 features \textbf{(b)}. The learned task-conditioned affordance model provides in-the-wild prediction for diverse fine-grained regions, which are used as observation space for manipulation policies \textbf{(c)}.
    }
    \label{fig:method}
    \vspace{-1.5em}
\end{figure*}

\section{Introduction}
\begingroup
  \setlength{\skip\footins}{1pt}    %
  \setlength{\footnotesep}{2pt}     %
  \let\thefootnote\relax
  \footnotetext{%
    Correspondence: Wenlong Huang
    (\href{mailto:wenlongh@stanford.edu}{wenlongh@stanford.edu})%
  }
\endgroup
Understanding the affordances of objects underpins a robot's capability to perform purposeful interactions in unstructured environments \cite{csahin2007afford,bahl2023affordances,hsu2023ditto}.
Given an open-ended task instruction specified in natural language, a robot must first identify the action possibilities afforded by the environment based on its visual perception. %
In particular, this understanding should extend beyond objects or object parts to encompass fine-grained details down to the level of pixels.
For instance, the robot might need to identify the exact grasp point on an unseen saucepot, the broom area to sweep trash, or the hanger's shoulders to hang clothes (Fig.~\ref{fig:teaser}).
While learning visual affordances from manually annotated datasets with closed vocabulary has been extensively explored in the literature~\cite{jamone2016affordances,yamanobe2017brief,hassanin2021visual,chen2023survey,liu2023survey,yang2023recent}, scaling affordance learning to open-world scenarios conditioned on free-form task instructions remains a long-standing challenge.

Vision-language models (VLMs) have demonstrated the ability to internalize world knowledge by pretraining on large-scale image-text datasets~\cite{openai2023gpt,team2023gemini}. Recent works also suggest that they encode affordance knowledge in the language space~\cite{achiam2023gpt} (e.g., ``handle should be grasped for opening drawers'').
However, the effective grounding of this knowledge in the continuous \emph{spatial} domain remains an open question.
In contrast, self-supervised vision models~\cite{caron2021emerging, oquab2023dinov2} provide general-purpose pixel-level features that capture low-level structures of objects. However, they are not conditioned on specific open-world task semantics, which is imperative for task-level generalization in robotic manipulation.

In this work, we introduce \textbf{\algname~(\algabrvname)}, a method that extracts affordance knowledge from foundation models and distills it into a task-conditioned affordance model, \emph{all without manual annotation}. Notably, \algabrvname leverages the complementary strengths of vision-language models and large vision models to automatically annotate a large-scale dataset with fine-grained \texttt{<instruction, visual affordance>} pairs. We then use the dataset to train a task-conditioned affordance model, by reusing and freezing the weights of DINOv2~\cite{oquab2023dinov2} and training only a lightweight task-conditioned decoder. We demonstrate a superior performance of the model evaluated \emph{zero-shot} on existing benchmarks~\cite{Luo2022arXiv}, along with exceptional generalization to a real-world robotic dataset~\cite{khazatsky2024droid} involving unseen objects in novel environments.

To translate these generalization properties into robust manipulation behaviors, we propose an imitation learning policy that uses \emph{affordance as the observation space}~\cite{jonschkowski2015learning} provided by the pre-trained \algabrvname. This approach sidesteps the common challenge of learning generalizable visual representations in vision-based manipulation on scarce interaction data and provides a manipulation-centric alternative to various pre-trained visual representations that are often tailored for vision tasks, such as CLIP~\cite{radford2021learning}.
Specifically, we demonstrate that the proposed framework possesses the unique advantage of generalizing to unseen environment configurations, object instances, object categories, and even novel task instructions, after training on as few as 10 demonstrations.

To summarize, our contributions are as follows:
1) We propose an unsupervised pipeline to automatically extract fine-grained affordance annotations using off-the-shelf vision-language models (VLMs) and large vision models (LVMs);
2) We scale the training of a task-conditioned affordance model that outperforms prior methods on existing benchmarks, despite evaluated zero-shot;
3) We show that using affordance as the observation space in an imitation learning policy enables generalization to unseen environments, object instances, object categories, and task instructions, while training with only a handful of demonstrations.

\section{Related Works}

\subsection{Learning and Discovering Visual Affordance for Robotics}
\emph{Affordance}~\cite{gibson1977theory} can be defined as action possibilities that are readily perceivable by an actor \cite{norman2013design}. This topic has two levels, namely learning and discovering affordance, and using affordance for downstream tasks \cite{yang2023recent, DBLP:journals/corr/abs-2104-00878}. These topics have been extensively studied in robotics and related fields as covered in several recent surveys \cite{jamone2016affordances,yamanobe2017brief,hassanin2021visual,chen2023survey,liu2023survey,yang2023recent}.
Affordance is typically expressed in perceptual space of the agent.
They differ in how the afforded actions are inferred: one can infer the action from probability maps (e.g., action possibility estimates), or by a direct mapping from the observations (e.g., keypoints or descriptors \cite{schmidt2016self,florence2018dense,manuelli2019kpam,kulkarni2019unsupervised,qin2020keto,sundaresan2020learning,manuelli2020keypoints,chen2021unsupervised,simeonov2022neural,simeonov2023se,vecerik2023robotap,chun2023local,wen2023any,bharadhwaj2024track2act,xu2021affordance,huang2024rekep}). Action possibilities are often represented as affordance maps, e.g., in the formats of probability distributions over image space \cite{myers2015affordance,kokic2017affordance,nguyen2017object,do2018affordancenet,chu2019toward,mandikal2021learning,hamalainen2019affordance,borja2022affordance,bahl2023affordances,bharadhwaj2023visual,srirama2024hrp} or continuous action possibilities \cite{moldovan2012learning,pohl2020affordance,zeng2022robotic,cai2019metagrasp,xu2021affordance,wu2020learning,yang2021learning,xu2021deep,mo2021where2act,ahn2022can,mees2022grounding,wu2023learning}, which have the same dimensions as the input image. Their values typically
indicate the likelihood of executing a certain action at each pixel location \cite{yang2023recent}. To learn a model that predicts affordance, deep learning-based methods are widely adopted \cite{chen2023survey}, which require a large amount of training data. For training, one can utilize supervised learning with existing datasets (e.g., \cite{myers2015affordance,Luo2022CVPR,Li2023,jian2023affordpose,deng20213d}) or self-supervised learning \cite{cai2019metagrasp,wu2020learning,yang2021learning,qin2020keto,turpin2021gift}.
While many works focus on developing models to be trained on existing datasets, our work uniquely investigates extracting affordance from large general-purpose, pre-trained models.

\subsection{Pre-trained Visual Representation for Manipulation}
An important application of~\algabrvname is visuomotor learning for robotic manipulation.
Specifically, we incorporate~\algabrvname as the observation space for robot policy, akin to related literature studying pre-trained visual representation for manipulation~\cite{shah2021rrl,pari2021surprising,parisi2022unsurprising,xiao2022masked,radosavovic2023real,ma2022vip,majumdar2023we,burns2023makes,yadav2023offline,nair2022r3m,karamcheti2023language,ma2023liv,khandelwal2022simple,cui2022can,shridhar2022cliport},
which can be coarsely categorized into those that are task-agnostic~\cite{shah2021rrl,pari2021surprising,parisi2022unsurprising,xiao2022masked,radosavovic2023real,ma2022vip,majumdar2023we,burns2023makes,yadav2023offline,lin2024spawnnet, wang2024gendp, jiang2024robotspretrainrobotsmanipulationcentric} and those that are task-conditioned~\cite{nair2022r3m,karamcheti2023language,ma2023liv,khandelwal2022simple,cui2022can,shridhar2022cliport, huang2025ottervisionlanguageactionmodeltextaware}.
We study the later setting, where visual representation differs depending on the agent's objective.
To learn the association between visual features and language features, previous work typically relies on a CLIP-like~\cite{radford2021learning} objective, which often exhibits ``bag-of-words'' behaviors that focus little on fine-grained visual details, as suggested by recent studies~\cite{tong2024eyes,thrush2022winoground,yuksekgonul2023and,hsieh2024sugarcrepe}.
In this work, we aim to address such limitations by proposing a data annotation pipeline that can effectively scale up the training of task-conditioned visual features \emph{without human annotations} while focusing on fine-grained predictions.

\subsection{Foundation Models for Robotics}
Leveraging foundation models for robotics is an active area of research~\cite{hu2023toward,firoozi2023foundation,kawaharazuka2024real,yang2023foundation},
with many works focusing on open-world reasoning and goal specification~\cite{huang2024copa,liu2024moka,nasiriany2024pivot,hu2023look,du2023video,hong20233d,chen2024spatialvlm,huang2023voxposer,brohan2023rt,gao2023physically,wang2024grounding,hsu2023ns3d,gao2024physically,yuan2024robopoint,duan2024manipulate}.
In this work, we are interested in acquiring general-purpose knowledge about affordance from existing foundation models through extraction and distillation, to obtain a model that maps task instructions to visual affordance.
To this end, we focus on VLMs that can perform visual question answering~\cite{radford2021learning,ramesh2021zero,li2022blip,achiam2023gpt,li2023blip,openai2023gpt} and self-supervised vision models~\cite{he2022masked,caron2020unsupervised,baevski2022data2vec,chen2020simple,caron2021emerging,oquab2023dinov2,darcet2023vision,wang2024segmentsupervision} that can provide fine-grained pixel-level features.
However, none of the aforementioned models directly supports the desired input-output mapping in this work.
As a result, the proposed~\algabrvname consists of a two-stage extraction and distillation process, with the critical insight being reformatting the visual affordance understanding as a visual question-answering problem.
Similar visual prompting techniques are also explored in prior work~\cite{nasiriany2024pivot,liu2024moka,huang2024copa,huang2024rekep}.
In comparison, in this work, we further distill the extracted knowledge into a specialized visual affordance model that is not only more efficient but also provides \emph{fine-grained continuous} predictions \emph{directly in visual space}.
Furthermore, we focus our study on its extensive utility in supporting generalization to various conditions in robotic manipulation.

\begin{figure*}[t]
    \centering
    \includegraphics[width=1.0\linewidth]{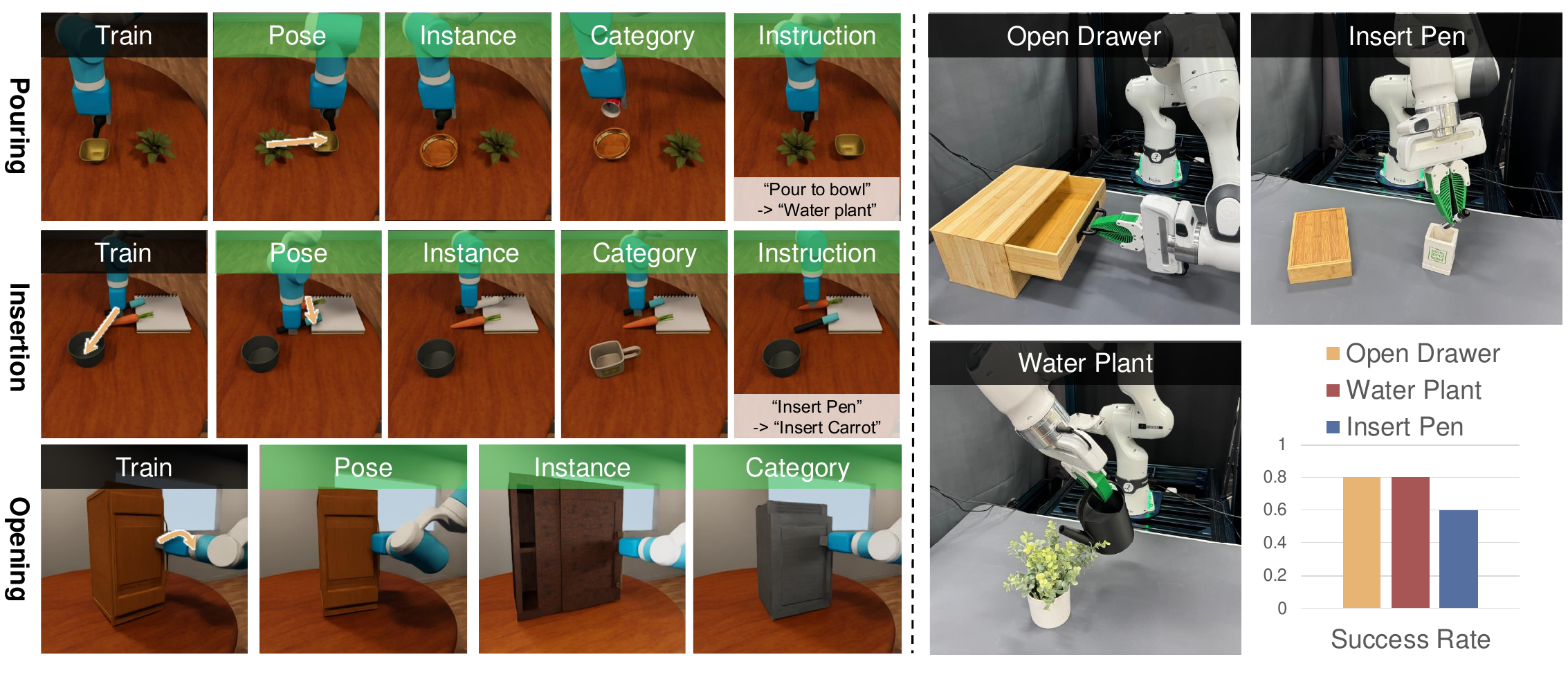}
    \vspace{-2em}
    \caption{Tasks for evaluating \algabrvname. Left: tasks in simulation along with different generalization requirements. Right: tasks in the real world and the corresponding success rate achieved by \algabrvname-based policies. }
    \vspace{-1.6em}
    \label{fig:policy}
\end{figure*}

\section{Method}
In this section, we discuss (A) how we extract affordance annotations from foundation models, (B) how we train a task-conditioned affordance model based on these annotations, and (C) how we leverage the learned affordance in imitation learning policies by using affordance as the observation space for generalization for robotic manipulation.

\subsection{Extracting Affordance Annotations} \label{sec:data-pipeline}
We are interested in visual affordance on pixel-level functional regions of objects, which we posit to be useful for downstream vision-based manipulation tasks.
However, manually labeling a large-scale affordance dataset is costly. Therefore, we want to extract affordance annotations from existing foundation models to construct a diverse dataset of the following triplets:
RGB images $I \in \mathbb{R}^{H \times W \times 3}$, free-form task instructions $\mathcal{T}$, and task-conditioned affordance map $A \in [0, 1]^{H \times W}$.

\textbf{Dataset.}
Although we are interested in obtaining a dataset of 2D annotations, we empirically find that the proposed pipeline performs significantly better when 3D consistency is enforced---a similar observation was also made in the recent investigation of pre-trained 2D visual features for manipulation and open-vocabulary 3D segmentation~\cite{wang2023d3fields,wang2024gendp, ma20243d}.
To this end, we focus on generating unsupervised affordance annotations in 2D from individually rendered 3D objects.
We will later discuss how we can train an affordance model that nevertheless generalizes to multi-object scenes even in the real world despite only being trained on single objects rendered in simulation.

We use a subset of the 3D assets from BEHAVIOR-1K~\cite{li2023behavior}, as the objects are tailored for the manipulation context. In total, the object database consists of 206 objects from 76 object categories, along with 667 task instructions.
Post paper acceptance, we additionally conduct a case study of scaling to more diverse object database, such as Objaverse-XL~\cite{objaverseXL}, which collectively amounts to more than 10,000 object-instruction pairs. Details are provided in Appendix.

An overview of our pipeline is shown in Fig.~\ref{fig:method}(a).
At a high level, we leverage LVMs to find fine-grained semantic regions for each object and VLMs to propose candidate task instructions relevant to each object. Then, we use VLMs to associate the regions and the task instructions. In the following, we introduce each component in order:

\textbf{Fine-Grained Region Proposal.}
For each 3D object, we first spawn it in an empty scene and render 14 views around the object to obtain $K$ RGB images $I_{i=1}^{K} \in \mathbb{R}^{H \times W \times 3}$ and its aggregated point cloud in world coordinates $P \in \mathbb{R}^{N \times 3}$.
For each \(I_i\), we extract pixel-wise features \(F_i \in \mathbb{R}^{H \times W \times d}\) from DINOv2.
We then adopt the procedure in~\cite{wang2023d3fields} to fuse the 2D features to obtain a global 3D feature field \(F_{\text{global}} \in \mathbb{R}^{N \times d}\) over the object point cloud $P$.
To obtain fine-grained semantic regions of objects, we first apply PCA on \(F_{\text{global}}\) to obtain \(F_{\text{reduced}} \in \mathbb{R}^{N \times 3}\), which is found to make features less sensitive to local texture.
Finally, we cluster the object points into \(M\) regions using the Euclidean distance on their features \(F_{\text{reduced}}\), where the clustering algorithm automatically identifies the number of clusters $M$.
We denote the region labels for all object points as $r_{n=1}^{N} \in \{1, ..., M\}$.

\textbf{Task Instruction Proposal.}
To identify the plausible task instructions for each object conditioning on the proposed regions, we perform visual prompting using a VLM, i.e., GPT-4o~\cite{openai2023gpt} in this work.
For each object, we first identify a natural-looking view by calculating the cosine similarity between the object category name and the corresponding RGB image under CLIP~\cite{radford2021learning} embedding.
We then visualize the proposed regions by assigning unique colors and overlay them on the original image.
We provide the overlaid image, original image, and object category name as input for the VLM and query it to propose a set of task instructions \(\{\mathcal{T}_1, ..., \mathcal{T}_J\}\) associated with this object. For instance, for a coffee mug, the VLM will propose task instructions such as ``rim of the coffee mug - region for drinking and pouring''.
The complete prompts can be found in the Appendix.

\textbf{Region and Instruction Mapping}
We then use a similar procedure to query the VLM to associate the task instructions it proposed with the most appropriate clustered region.
As a result, we obtain a mapping from each task instruction to exactly one region on the object.
Eventually, we want to create a continuous affordance map \(A \in [0, 1]^{H \times W}\) because we believe that affordance is fundamentally continuous rather than binary. Certain regions are more closely associated to the specified task (e.g. middle of a handle for grasping) than others (e.g. the tip of the handle), which is better captured by a continuous formulation. To do so, for each region $r$ identified by the VLM, we first average the features of the corresponding points to obtain a reference feature $f_\text{ref} \in \mathbb{R}^d$.
Then we compute the cosine similarity score between \(f_\text{ref}\) and \(F_{\text{global}}\) to obtain a $[0, 1]$ similarity score for each 3D point.
Finally, we project object points along with their scores to each camera view to obtain the final affordance map \(A \in [0, 1]^{H \times W}\).
This operation converts \emph{discrete} decisions produced by VLM to \emph{continuous} pixel-level values; intuitively, it can be interpreted as per-pixel likelihood of whether it ``affords'' the given task.
This is in contrast to many prior works~\cite{8315047} that consider image-level distributions (i.e., all pixels sum to one) where affordance magnitude is normalized by afforded region size.
In summary, the pipeline produces a dataset with triplets of ($I$, $\mathcal{T}$, $A$).

\subsection{Learning Task-conditioned Affordance Model} \label{sec:vision-model-training}
To train an affordance model that generalizes to real-world multi-object scenes using only synthetic single-object data, we leverage the pre-trained DINOv2~\cite{oquab2023dinov2} by freezing its weights and only training a lightweight language-conditioned module on top.
Specifically, we first obtain the language embedding \( e_{\mathcal{T}} \) for each entry in the dataset using OpenAI APIs.
To condition the features from DINOv2 on the language embeddings, we use FiLM layers~\cite{perez2018film} that take in the language embedding $e_{\mathcal{T}}$ as well as the pixel-space features $X \in \mathbb{R}^{H \times W \times C_{\text{in}}}$ and output $X' \in \mathbb{R}^{H \times W \times C_{\text{out}}}$, where $C_{\text{in}}$ and $C_{\text{out}}$ are input and output channel dimensions, respectively.
We use 3 FiLM layers with output channels \( [256, 64, 1] \), which produce logits at each pixel location as the final output $\hat{A} \in [0, 1]^{H \times W}$.
We note that the learned transformation for each channel by FiLM is agnostic to pixel location, which is suitable for our intent to build an association between the DINOv2 feature and task instructions.
We use binary cross-entropy as the loss function, computed between the ground truth affordance map $A$ and the predicted logits of the affordance map $\hat{A}$.
We term the learned affordance model as~\algabrvname.

\subsection{Policy Learning with Affordance as Observation Space}
\label{subsec:policy_method}
\algabrvname can be naturally integrated into existing vision-based policy architectures for manipulation as an encoder for the visual input.
Effectively, instead of learning a \emph{task-agnostic} visual representation as in most existing policy architectures, \algabrvname serves as fine-grained visual attention for the policy that contains prior knowledge \emph{conditioned on tasks at hand}.
To investigate its capability, we integrate~\algabrvname with a multi-view transformer policy adopted from RVT~\cite{goyal2023rvt,goyal2024rvt}.
We assume access to detailed language instructions (e.g., ``grasp watering can'', ``align spout'', ``water plant'').
Using the given instruction, we first predict the affordance map for each view $\mathbb{R}^{H \times W}$.
Then, we follow RVT to augment each view with the corresponding depth value and the \((x, y, z)\) coordinates of points in the world frame, as well as a global proprioception vector.
The policy outputs a 7-dimension action that includes a 6-DoF end-effector pose and a binary gripper action.
We train the policy using imitation learning and focus our investigation on its generalization capabilities.
Even though we do not finetune the affordance model in policy training, we can effectively train the policy using only a handful of demonstrations while exhibiting generalization capabilities to various conditions leveraging~\algabrvname.

\section{Experiments}

We seek to answer the following research questions:
(A) Despite only being trained on rendered 3D objects, can \algabrvname generalize to real-world scenes from existing robotic datasets in affordance prediction, and how does it compare to prior methods on visual affordance benchmarks?
(B) Using \algabrvname as observation space, what generalization properties does a visuomotor policy have compared to other pre-trained representations?
(C) How well does an \algabrvname-based policy perform in real-world manipulation tasks?

\subsection{Task-Conditioned Affordance Prediction}\label{sec:vision}
In this section, we focus on how \algabrvname performs on task-conditioned visual affordance prediction.
Note that we train a \emph{single} \algabrvname only on rendered 3D objects, which is used to perform evaluations across all settings discussed below.

\textbf{\algabrvname generalizes to novel instances, categories, and instructions on rendered objects.}
We first perform a sanity check on how well \algabrvname performs within the same domain of simulator rendered, single object images. We construct four evaluation sets of image-text pairs that contain, respectively, 1) training data, 2) novel object instances, 3) novel categories, and 4) novel instructions.
We use Amazon MTurk to obtain the ground truth for evaluations, with details in the Appendix.
Based on recent study~\cite{8315047}, we use the Area Under ROC Curve (AUC) as the metric for evaluation, as it evaluates the predicted affordance map as a per-pixel classifier of the ground-truth mask~\cite{8315047}, closest to our interpretation.
Evaluated on 100 \textless instruction, visual affordance\textgreater~pairs per setting, \algabrvname achieved an AUC score of at least 0.92 across all four settings, indicating its strong generalization capability, as well as the consistency between \algabrvname and human predictions. 

\textbf{Leveraging pre-trained features, \algabrvname can seamlessly generalize to real-world robotic scenes.}
To create an affordance prediction evaluation set tailored for manipulation, we investigate \algabrvname on a subset of DROID~\cite{khazatsky2024droid}, a real-world robotic dataset containing trajectories of robots performing manipulation tasks in diverse, in-the-wild scenes.
Specifically, we select task episodes from DROID that involve interaction with specific, fine-grained object regions, rather than tasks with ambiguous instructions (e.g., ``pick up the colored cube,'' where any part of the object can be manipulated).
We extend the built-in task descriptions with additional text details at the same level of granularity as those in training.
For example, the task ``pick up the lid and put it on the pot'' is broken down into ``pick up the lid'' and ``align with the rim of the pot''.
For each episode, we capture the first frame from two table-mounted third-person cameras.
We center-crop the images and filter out those where the key objects are not clearly visible (e.g., due to occlusion).
We follow the same procedure as in the previous section to obtain the ground-truth labels from Amazon MTurk.

\begin{figure}[t]
    \centering
    \includegraphics[width=1.0\linewidth]{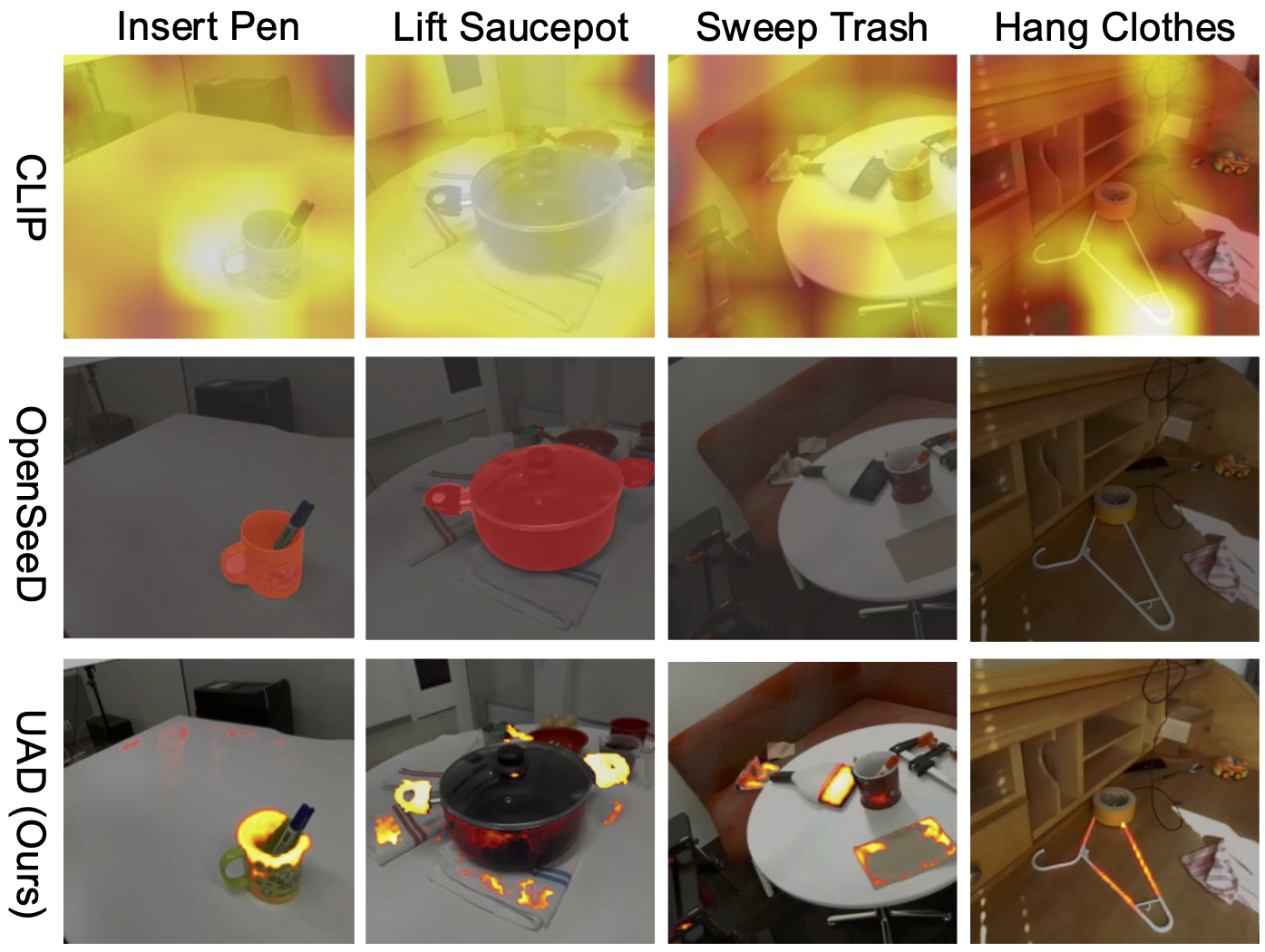}
    \caption{Task-conditioned affordance prediction results on the DROID dataset. Average AUC scores (evaluated on the entire dataset): 0.500 (CLIP), 0.836 (OpenSeeD), 0.840 (Ours).}
    \label{fig:droid}
    \vspace{-1.5em}
\end{figure}

We compare~\algabrvname to CLIP~\cite{radford2021learning} and OpenSeeD~\cite{zhang2023simple}, an open-vocabulary segmentation model.
Results are shown in Fig.~\ref{fig:droid}. Despite only trained on rendered single objects, by leveraging DINOv2~\cite{oquab2023dinov2} as backbone, \algabrvname can generalize to in-the-wild, multi-object, often even cluttered scenes.
Notably, compared to CLIP, \algabrvname provides much more fine-grained and robust features, often focused specifically on potential regions of interaction.
Compared to open-vocabulary segmentation, which outputs binary segmentation, \algabrvname produces a continuous representation.
Notably, even when target objects or parts are small, \algabrvname can consistently produce per-pixel, continuous prediction, whereas this is observed to be a typical failure case for segmentation models.

\begin{figure}[t]
    \centering
    \includegraphics[width=0.95\linewidth]{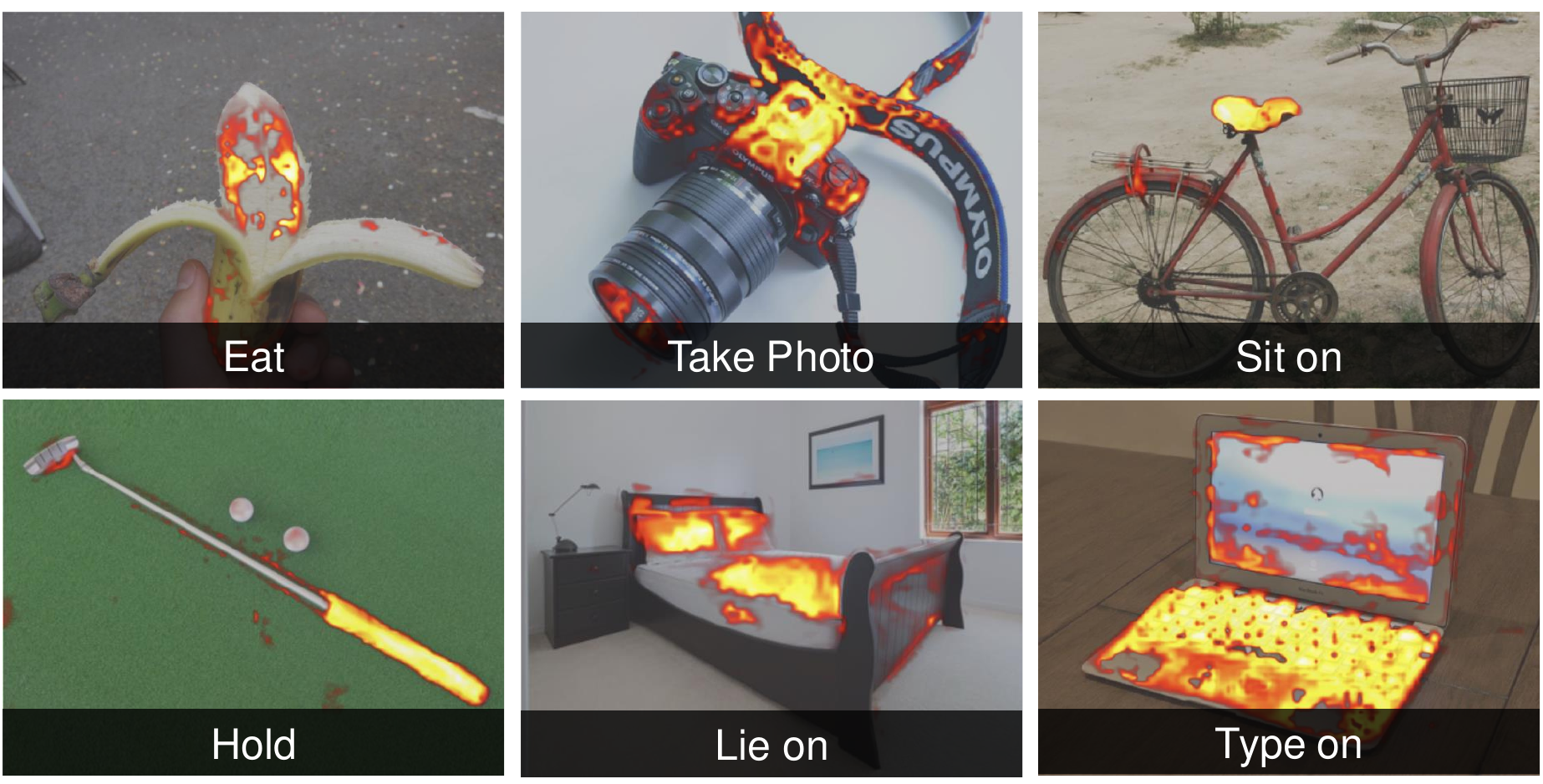}
    \caption{Zero-shot generalization to affordance predictions in human activities from AGD20K.}
    \label{fig:agd}
    \vspace{-1em}
\end{figure}

\textbf{\algabrvname performs competitively on human activity affordances even though certain activities or objects are entirely unseen.}
Motivated by the promising generalization of \algabrvname, we are also interested in investigating how \algabrvname may perform in existing affordance prediction benchmark focused on human activities, such as AGD20K~\cite{Luo2022CVPR}.
The results are shown in Fig.~\ref{fig:agd} and Tab.~\ref{tab:agd_quant}.

\begin{table}[!htb]
    \centering
    \begin{tabular}{@{}lcccc@{}}
    \toprule
    Methods           & KLD (↓) & SIM (↑) & NSS (↑) & NSS-0.5 (↑) \\ \midrule
    Cross-View-AG \cite{Luo2022CVPR}     & 1.787 & 0.285 & 0.829 & - \\
    LOCATE \cite{li2023locate}            & \textbf{1.405} & 0.372 & \textbf{1.157} & 1.723 \\
    3DOI \cite{Qian2023}              & 3.565 & 0.227 & 0.657 & - \\
    AffordanceLLM~\cite{qian2024affordancellm}      & 1.463 & 0.377 & 1.070 & - \\\midrule
    \algabrvname (Ours)               & 1.878 & \textbf{0.407} & 1.092 & \textbf{2.050} \\
    \bottomrule
    \end{tabular}
    \caption{Evaluation results on AGD20K Unseen test split.}
    \label{tab:agd_quant}
    \vspace{-2.5em}
\end{table}

For evaluation, since \algabrvname conditions on free-form language and AGD20K contains only a list of pre-defined actions, we format the instructions as ``region to \textless action\textgreater~the \textless object\textgreater''.
We follow the evaluation procedure defined in AGD20K \cite{Luo2022CVPR} and evaluate \algabrvname on the Unseen test split using the same metrics as in previous work, namely KL Divergence (KLD), Similarity Metric (SIM), and Normalized Scanpath Saliency (NSS).
We additionally report NSS-0.5, which uses a stricter threshold (0.5 vs. standard 0.1) to focus evaluation on the most salient ground truth regions. See Appendix ~\ref{sec:vision} for details. Despite not being trained on AGD20K data, \algabrvname performs competitively, achieving the highest SIM score and outperforming LOCATE on NSS-0.5. The higher KLD stems from \algabrvname's fine-grained predictions compared to AGD20K's diffuse ground truth distributions. 
Interestingly, \algabrvname can even generalize to a number of human activities that involve objects and task instructions completely out of distribution from our training set, such as ``eating bananas'', ``taking photos'', ``sitting on bicycles'', ``holding golf clubs'', ``lying on bed'', and ``typing on computers'', as shown in Fig.~\ref{fig:agd}.

\subsection{Policy Learning in Simulation}\label{sec:sim}

Using~\algabrvname as observation space (Fig.~\ref{fig:method}(c)), we evaluate the generalization properties of a transformer-based policy learned via imitation learning.
We perform our experiments in OmniGibson~\cite{li2023behavior}, equipped with photo-realistic rendering and a variety of everyday objects.

We select three tasks that require fine-grained reasoning of object affordance, \textit{Pouring}, \textit{Opening}, and \textit{Insertion}, each with varying objects designed to assess the generalization performance of the policy (Fig.~\ref{fig:policy}).
For each task, we train a policy on 10 scripted demonstrations with randomization in object poses.
We evaluate the trained policy against four generalization settings: new object poses, instances/models, categories, and task instructions (Fig.~\ref{fig:policy}).
To better attribute generalization capabilities solely brought by affordance prediction, for novel categories, we choose objects with similar functional structure---for example, one needs to grasp the lower body to lift up both the beer bottle and Coke can.
Since \algabrvname focuses on generalization in task-conditioned visual prediction, when designing evaluation to novel instructions, we focus on scenarios where correct identifying affordance would lead to successful completion of the task. For instance, for both pouring fluid and watering plants, the robot needs to approach the correct region near the target object (bowl and pot plant, respectively) and tilt the fluid container.

We compare against baseline policies that use RGB images or other pre-trained visual representations as observations, including DINOv2~\cite{oquab2023dinov2}, CLIP~\cite{radford2021learning}, and Voltron~\cite{karamcheti2023language}. %
Additional setup details can be found in the Appendix.
The results are shown in Fig.~\ref{fig:main_results}. Each bar represents the average success rate across all three tasks, with each task-generalization setup combination being evaluated over 15 trials.
In general, although trained on only a handful of demonstrations, the \algabrvname-based policy demonstrates promising generalization capabilities in all settings evaluated.
We summarize our main takeaways below:
\begin{compactitem}
\item  \algabrvname is robust against variations in object appearance, e.g. successfully manipulating white markers in \textit{Insertion} tasks despite training only on black ones..
\item \algabrvname is particularly advantageous on tasks requiring fine-grained visual perception,
allowing it to outperform baselines on the \textit{Opening} task involving detecting grasp points on thin handles of drawers.
\item As \algabrvname is conditioned on task instructions while offering precise affordance prediction, 
it can also generalize to variations in instructions that control the objects of interaction via natural language.
\end{compactitem}

\begin{figure}[t]
    \centering
    \includegraphics[width=1.0\linewidth]{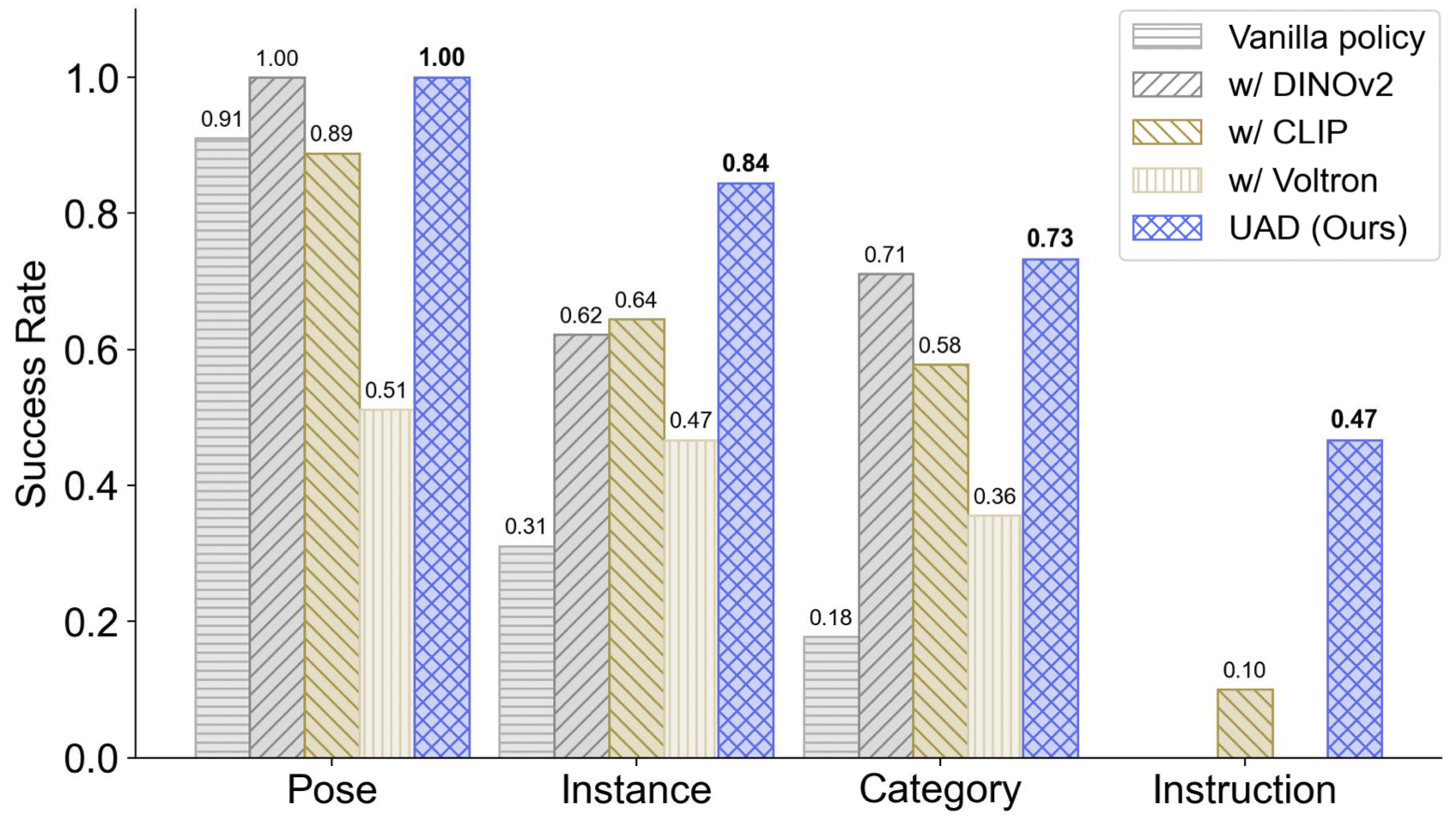}
    \caption{Generalization performance of \algabrvname in three simulation tasks. \algabrvname shows better generalization capabilities compared to the baselines.}
    \label{fig:main_results}
    \vspace{-1em}
\end{figure}

\subsection{Policy Learning in the Real World} \label{sec:real}
To demonstrate that \algabrvname-based policies can solve real-world tasks, we further evaluate models on three robotic manipulation tasks as shown in Fig.~\ref{fig:policy}.
We use a Franka Emika Panda robot with a tabletop setup (more details about hardware setup can be found in the Appendix), with two RGB-D cameras mounted on the opposite sides of the workspace.
Following the policy learning setup in Sec.~\ref{sec:sim}, we train a policy for each task using 10 human demonstrations collected using kinesthetic teaching.
Compared to the simulation setup, the real-world environment poses additional challenges, such as identifying affordances for visually diverse real-world objects, as well as subjecting to additional noise introduced by various components in the real-world system stack, including RGB-D cameras and low-level controllers.
The results are shown in Fig.~\ref{fig:policy}, with success rates averaged across 10 trials for each task. Overall, the \algabrvname-based policy can perform real-world manipulation tasks with an average success rate of $73\%$. The predicted affordance maps contain fine-grained details that allow for precise 6-DoF manipulation, such as inserting a pen and opening a drawer.

\section{Conclusion \& Limitations}
Learning and discovering object affordance is an important step toward generalizable robotic manipulation. We propose Unsupervised Affordance Distillation (\algabrvname), a novel method that distills affordance knowledge from foundation models into a task-conditioned affordance model, without relying on manually annotated datasets. Our model achieves competitive performance on existing affordance prediction benchmarks and demonstrates strong generalization capabilities in real-world robotic tasks. 

Despite the promising findings, a few limitations remain.
First, \algabrvname focuses on extracting visual affordance from foundation models. Although we observe promising generalization when using it as the observation space for imitation learning policy, it does not immediately provide generalization at the motion level.
Second, we only consider the interpretation of affordance for a single static frame. However, manipulation tasks are typically concerned with multi-step visual understanding and behaviors.
Third, the extracted training dataset contains only single objects renderings -- extending the annotations to real-world multi-object images may enable better grounding of world knowledge in foundation models to continuous spatial domains.

\section*{ACKNOWLEDGMENT}
This work is in part supported by NSF RI \#2211258 and \#2338203, ONR N00014-23-1-2355, ONR MURI N00014-22-1-2740, and ONR MURI N00014-24-1-2748.

\newpage

\bibliographystyle{IEEEtran}
\bibliography{IEEEabrv,main}

\clearpage
\appendix

\subsection{Details on Unsupervised Affordance Annotation Extraction Pipeline (Sec \ref{sec:data-pipeline})}

\textbf{Implementation Details on Fine-Grained Region Proposal.} In this section we discuss additional implementation details to obtain candidate regions. 

From multi-view RGBD renderings of an asset, we obtain aggregated point cloud $P \in \mathbb{R}^{N \times 3}$ by projecting fore-ground pixels (we obtain this mask from the rendering simulator) of each view to world frame and uniformly down-sample the point cloud. Then for each RGB image, we extract patch-wise features from DINOv2 with registers (ViT-L14)~\cite{oquab2023dinov2} and perform bilinear interpolation to upsample the features to original image size. 

To fuse the DINOv2 features from all views to $P$, we adapt the following procedure from \cite{wang2023d3fields}: for each point $p \in P$, we compute it's corresponding pixel on each camera view. We consider it to be visible in a camera view if the projection depth is close to the depth image reading at that pixel by a small threshold. The fused feature for $p$ is the average of DINOv2 features on it's corresponding pixel across all the views $p$ is visible.

With the procedure above we obtain global feature field \(F_{\text{global}} \in \mathbb{R}^{N \times d}\), and then we apply PCA to obtain \(F_{\text{reduced}} \in \mathbb{R}^{N \times 3}\) to mitigates affect of local texture or appearance on cluster result. 

We group object points into candidate regions by running clustering algorithm on \(F_{\text{reduced}}\). Since our data processing pipeline handles over-segmentation better than under-segmentation (reasons established in next su-bsection), we first run Mean Shift on the features, and if it found less than 5 clusters over the object, we re-run the $k$-means to find 5 clusters. Specifically, for articulated objects such as cabinets, we obtain the per-link mask from the rendering process and run the aforementioned clustering pipeline for each link individually. This allows us to find finer object regions such as drawer knobs. 

\textbf{Implementation Details on Task Instruction Proposal and Region-Instruction Mapping.} After we obtain the region labels for all object points, we visualize the clustering on the view we selected with the following procedure: for each foreground pixel in that view, we compute the corresponding 3D point with using the depth map. Then we find its closest neighbor $p$ in the aggregated pointcloud $P$, and use the label $r_p$ as the region label for this pixel. 

We assign a unique color to all pixels within each cluster and overlay this on the original RGB image. We input the cluster visualization with the original image to the VLM prompt below. The prompt contains only generic instructions and a few text-based examples to illustrate expected output and format. VLM is queried to propose a set of task instructions \(\{\mathcal{T}_1, ..., \mathcal{T}_J\}\) closely related to the object, and associate each instruction with a single candidate region. We use GPT-4o~\cite{openai2023gpt} for all our experiments.

To convert the instruction-region matching to continuous affordance map \(A \in [0, 1]^{H \times W}\), we average the features for points in the corresponding region and calculate cosine similarity score of this reference feature and \(F_{global}\). We project this to each camera view following the same procedure as we visualize cluster results, i.e. for each pixel, find its corresponding 3D point's nearest neighbor in \(P\) and assign that point's value. All values below 0 are set as 0 to ensure the correct value range of the obtained affordance map.

We found over-segmentation by the clustering step is preferred over under-segmentation. When a object part is over-segmented, empirically GPT-4o is still capable for correctly associating the instruction with one of the regions, and through the cosine similarity calculation, the other not selected regions within the same part is likely still computed to have high cosine similarity to the reference feature. On the other hand, under-segmentation could cause the affordance map to highlight regions that are not most closely related to the instruction, which is not desired for our purpose of finding fine-grained affordance. 

\begin{figure*}[h]
\begin{lstlisting}
## System Prompt

Given (a) the category of an object, (b) a reference image of the original image, (c) an image visualizing clustering of the object into a few regions, each indicated by a distinct color, and (d) a related list of used colors, please:
1. Identify specific regions of the object that serve different purposes in various manipulation tasks. 
- Focus on crucial parts and offer detailed and fine-grained descriptions of the regions of interest.
- For each identified region, provide both a Region Description and a Region for action xxx. For example, "handle of plastic bag -- region for agent to hold and lift the bag." Use double quotes only to represent a string element. 
- Avoid using multiple single quotes for an element. For example, instead of "adjusting the lamp's position", use "adjusting the lamp's position." Avoid trivial regions like: power cord, power plug, seal, small edges, small corners, different sides of the walls or body, interior base, or exterior edge.
2. Match the colored region in (c) with the proposed task in step 1, considering the functionality and the granularity of the task. The requirements are as follows:
- Compare the original image to the proposals to find the colored region that matches the description, considering the context provided by the explanation. For example, if a given proposal image indicates that the red region covers the handle, and the description mentions a task related to the handle, you should identify the answer as "Red."
- When the described region is clustered into more than one cluster on the image, pick the cluster that is most appropriate according to the explanation provided in the description.
- If the described region is within a cluster but still contains other parts of the object, you should still select the cluster.
- Consider the reason/explanation to make the final decision, and provide your best guess.
If you cannot identify the counterpart on the given image, give your best guess. Do not say you cannot identify something.

## User Prompt

The first image is the original image of the object, and the second image shows the clustering of regions in colors.
This is the color list: {CLUSTER_COLORS}, and this is the object category:{OBJ_CATEGORY}.
I need you to propose the task-guided fine-grained region description and match region description with the one most appropriate color in the images.
Output format: Start with the word "ANSWER: ", followed by a dict, where each key-value pair is in the format of "region description -- region for xxx" : "Color", separated by commas. Specifically, the content after "ANSWER: " should be parseable with Python's ast.literal_eval() and nothing else. All elements in the dict keys or values should be enclosed by double quotes only. The color could only be one color, with the first letter of the color name capitalized and the rest in lowercase.
\end{lstlisting}
\label{listing:vlm-prompt}
\end{figure*}

\section{Details on Task-conditioned Affordance Model Training (Sec \ref{sec:vision-model-training})} 

We obtain triplets of RGB object image, task instruction, and affordance map ($I$, $\mathcal{T}$, $A$), from our data extraction pipeline. We further process the affordance map by setting all values below a threshold 0.5 to 0, with the purpose to create a ground-truth map more focused on the most relevant regions. We then apply a Gaussian blur to $A$ with kernel size = 3, to accommodate for the boundaries created by the previous thresholding process and improve training stability. 

Our model contains 3 FiLM-conditioned convolution layers with output channels \( [256, 64, 1] \). We use linear layers to predict channel transformations from language embeddings. We initialize the linear layers with weights to be 1 and bias to be 0, adapted from implementation in RT-1~\cite{brohan2022rt}. We use Adam optimizer with a learning rate of 0.001. We train our model for 30 epochs with a batch size 8, which takes approximately 12 hours on a single NVIDIA A6000 GPU. 

\subsection{Details on Task-conditioned Affordance Prediction (Sec \ref{sec:vision})}

\begin{figure*}[h]
\centering
\includegraphics[width=0.8\textwidth]{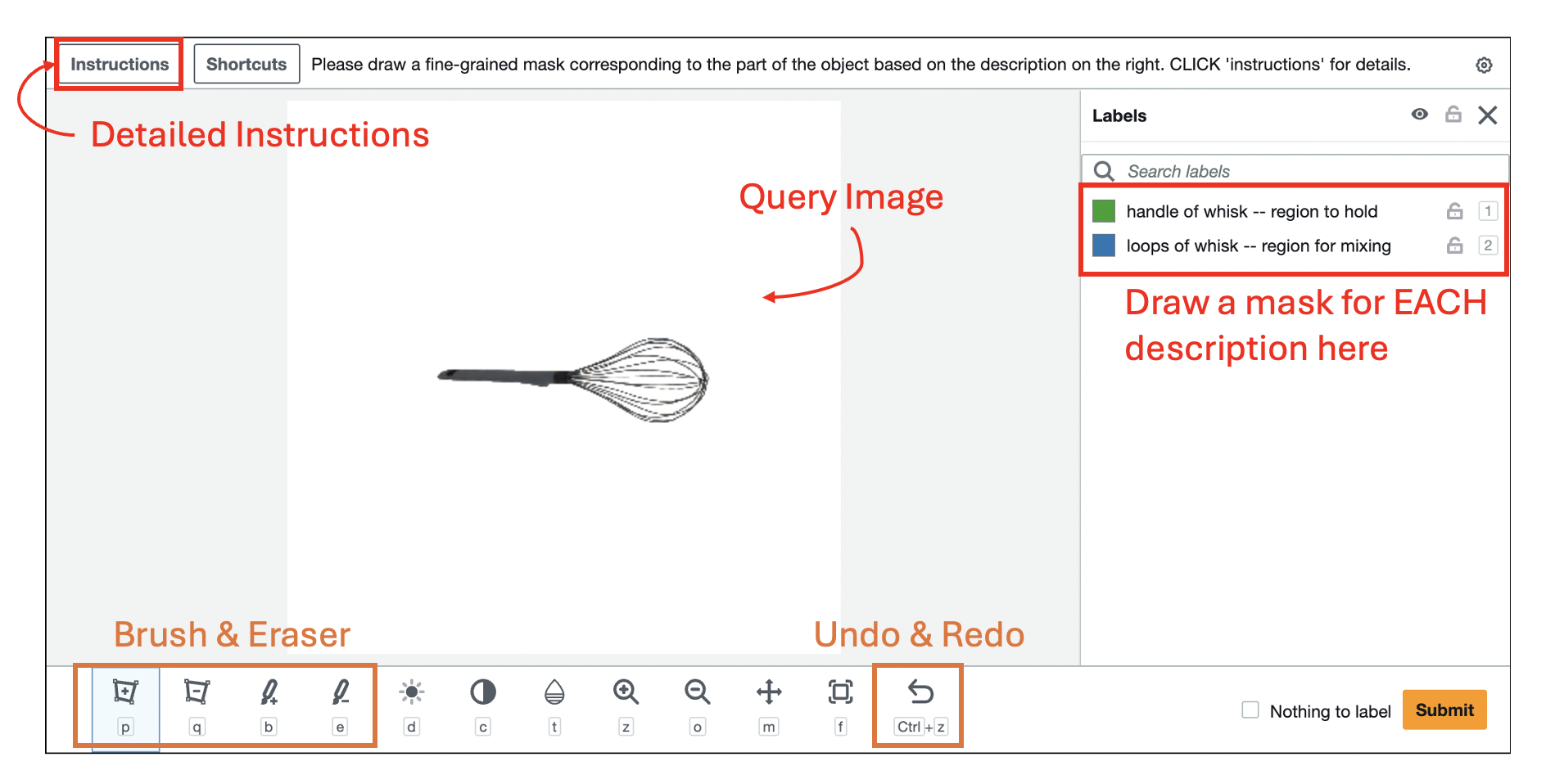}
\caption{Amazon MTurk Annotation Interface.}
\label{fig:mturk-interface}
\end{figure*}

\textbf{Mturk Interface.}
Fig \ref{fig:mturk-interface} is the Amazon MTurk interface we use to collect human affordance annotations on our valuation sets and DROID images. 

\textbf{MTurk Task Assignments and Label Post-processing.}
To obtain the ground-truth for evaluations, we prompt the image and corresponding text to Amazon MTurk workers and ask them to draw a fine-grained mask on the image for the region they believe corresponds to the text. On average, worker complete labelling assignment for each image in 40 seconds, for which they are compensated for 0.6 dollars. 

For each (text, image) query pair, we collect annotations from 7 MTurk workers and apply a pixel-wise voting scheme. A pixel is marked as part of the ground truth mask (value 1) if more than three workers label it accordingly. 

\textbf{Details for Evaluation on AGD20K Dataset -- Evaluation setting and Processing.}
We evaluate our model on the Unseen test split of AGD20K and compare with the baseline performance reported in~\cite{qian2024affordancellm}. To avoid numerical instability in KLD computation as we consider per-pixel affordance instead of per-image affordance as in other works, we post-process our model's prediction by clipping per-pixel prediction to $[\epsilon, 1-\epsilon]$ before normalization. 

We format instructions as "region to \textit{action} the \textit{object}" to match AGD20K's action vocabulary. However, AGD20K's ground truth annotations contain ambiguities in some action-region mappings or select one specific region when multiple valid options exist. For instance: 
\begin{compactitem}
    \item For "hit" with an axe, the ground truth highlights the handle (for gripping during hitting) rather than the head (which does the hitting).
    \item For "hold" with a cup, the ground truth exclusively marks the handle, though the cup body is also commonly used for stable holding.
\end{compactitem}
Since UAD was not trained on AGD20K data, it lacks knowledge of these dataset-specific conventions. To ensure fair evaluation, we disambiguated these instructions to match the ground truth interpretations:
\begin{compactitem}
    \item hit - axe → handle of axe to hold during hitting
    \item ride - bicycle → region to sit on and push the bicycle
    \item pour - cup → handle of the cup to hold while pouring
    \item wash - cup → rim of the cup to wash
    \item hold - cup → handle to hold the cup
\end{compactitem}

\begin{figure}[h]
    \centering
    \vspace{-0.8em}
    \includegraphics[width=\linewidth]{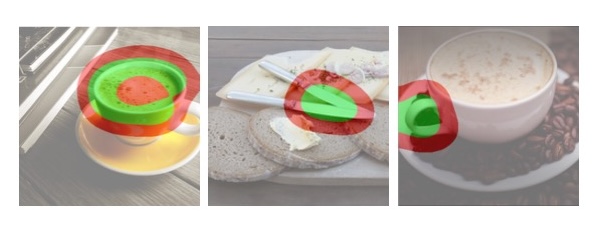}
    \caption{Comparison of ground truth fixation regions at different NSS thresholds. Green regions indicate pixels with GT $> 0.5$ (used for NSS-0.5), while red regions show pixels with GT values between 0.1 and 0.5 (additionally included in standard NSS). }
    \label{fig:nss_threshold}
    \vspace{-0.8em}
\end{figure}

\textbf{Details for Evaluation on AGD20K Dataset -- NSS-0.5 metric.} The NSS metric evaluates model predictions at``fixation pixels'', defined as locations where ground truth affordance values exceed a specified threshold. It computes the average prediction at these pixels, with all pixels weighted equally. The standard evaluation uses a relatively low threshold (0.1). However, the AGD20K ground truth is generated by computing Gaussian mixtures centered on manually annotated keypoints, causing the affordance signal to diffuse beyond actual object boundaries into background regions. As visualized in Fig.\ref{fig:nss_threshold}, the green areas represent pixels with GT $>$ 0.5, while the red areas show pixels with GT values between 0.1 and 0.5, which often extend into the background and may not correspond to actual object affordances. Under this original evaluation protocol (threshold=0.1), models are penalized for predicting low affordance in these peripheral regions (red), even though they may not correspond to actual object affordances. By using threshold=0.5, NSS-0.5 focuses evaluation on the core affordance regions (green) that are more strongly associated with the query. Our results show that UAD outperforms LOCATE—the previous state-of-the-art on standard NSS—on this more stringent metric, demonstrating our model's ability to identify salient affordance regions.

\textbf{Note on KLD scores.} The KL Divergence metric heavily penalizes predictions that approach zero where ground truth values are positive. As shown in our threshold visualization, the AGD20K ground truth extends into background regions due to its Gaussian mixture construction. Since \algabrvname's predictions are focused on foreground objects and predict near-zero values in these peripheral areas where the ground truth retains small positive values, this partially contributes to \algabrvname's higher KLD score.

\textbf{Details for Evaluation on DROID Images.}
We implement the baselines as follows:
\begin{compactitem}
    \item CLIP: we obtain the per-patch visual feature of query image and use bilinear interpolation to original image height and width. We compute the per-pixel cosine similarity with the text feature of query instruction and clip the minimum value to be 0.
    \item OpenSeeD: Since OpenSeeD is an open-vocabulary segmentation model, we use the same query instructions as for other methods to query the predicted mask. We set all pixels in the predicted object mask to 1 and 0 otherwise. The prediction image is all 0 if no mask is found. 
\end{compactitem}

\subsection{Policy Learning in Simulation Details (Sec \ref{sec:sim})}

\textbf{Environment Setup.}
Our simulation environment in OmniGibson contains one Fetch robot, for which we use operational space controller for the end-effector pose, multi-finger gripper controller for the gripper, and kept location of the robot base fixed. Grasping is physically simulated for all tasks. 

We use a key-frame based policy for both demonstrations and learnt policies.
Key-frames are commonly used by prior works~\cite{shridhar2022peract,goyal2023rvt,goyal2024rvt,james2022q,james2022coarse} as ``important or bottleneck steps of gripper during task execution". To execute an action $<$end-effector pose, gripper action$>$, we first command the end-effector controller of an interpolated trajectory from current to target pose, then execute the gripper action afterwards. 

We use 3 cameras at the front, left, and right of the workspace for \textit{Pouring} and \textit{Inserting}. We use 2 cameras on both sides of the robot for \textit{Opening} as the articulated objects are typically large in size and would occlude the other cameras.

\textbf{Tasks.}
Below we discuss the details of environment setup, scripted policy steps, success criteria, and evaluation generalization setting for each task. 

\textbf{\textit{Pouring}} 
The environment includes a beer bottle, a bowl, and a pot plant. The scripted policy involves four key-frames: reaching a pre-grasp pose next to the bottle, grasping the bottle, lifting and moving it next to the bowl, and tilting it to pour into the bowl. Success is defined by the alignment and tilting of the bottle's opening directly over the bowl.

At training time, object poses are randomized within a [±5cm, ±3cm, 0] range, with the bowl and the pot plant positions randomly swapped. This randomization is maintained during evaluations to test the system with varied object poses.

Different object models for the beer bottle and bowl are used for the novel object instance evaluation. The beer bottle is replaced with a Coke can in the novel object category evaluation. For the novel instruction, the task is changed to watering the pot plant.

\textbf{\textit{Opening}} 
The task environment features a cabinet with a revolute door. The task sequence includes two steps: reaching and grasping the cabinet door handle, followed by pulling it open. The task is considered successful if the door opens to at least 45 degrees.

During training, the position and orientation of the cabinet are randomized within a range of [±5cm, ±5cm, 0] for position, and ±15 degrees around the z-axis for rotation. This randomization is also applied during evaluations to assess performance with varied object poses.

For the novel object instance evaluation, a different cabinet model is used. A small refrigerator substitutes the cabinet in the novel object category setting, testing adaptability to different objects. Novel instruction scenarios are not evaluated for this task.

\textbf{\textit{Insertion}}
The environment contains a marker, a carrot, and a pencil holder. The task involves two key steps: picking up the marker and positioning it directly above the pencil holder’s opening in an upright orientation. The task is considered successful if the marker is in the holder.

During training, the positions of the pen and carrot are randomized within ±1.5 cm in the x-direction, and the pencil holder is adjusted within ±3 cm in both x and y directions. The pen and carrot positions are also randomly swapped. The same randomization parameters are used during evaluation.

A different marker model, varying in color and size, is used for evaluating a new object instance. The pencil holder is replaced with a coffee cup for the novel object category evaluation. The task of inserting the pen is changed to inserting the carrot for the novel instruction evaluation.

\textbf{Details on Baseline Visual Representations.} Herein we introduce our implementation for each baseline visual representations. 
\begin{compactitem}
    \item Vanilla policy: original rgb observation from each camera.
    \item w/ DINOv2: we first obtain per-pixel DINOv2 features for the rgb image of each camera. Then, we have a trainable 1D convolution layer with kernel size of 1 to reduce the number of channels to 3. 
    \item w/ CLIP: we obtain CLIP text embedding for each detailed instruction for the task. Then we calculate the cosine similarity against per-pixel CLIP visual embedding of each camera observation. 
    \item w/ Voltron \cite{karamcheti2023language}: we load a frozen Voltron (V-cond) model and obtain the visual embedding conditioned on task description. We interpolate the per-patch embedding to pixel space, and use a trainable 1D convolution layer with kernel size 1 to reduce the number of channels from 384 to 3. We have also experimented with using a trainable multi-head attention pooling layer for feature extraction, as suggested by the original paper, yet haven't observed improved performance.
\end{compactitem} 

\textbf{Training Details}
We trained each policy for 4000 epochs. Training with batch size of 3 on a single NVIDIA A40 GPU takes approximately 16 hours for \textit{Pouring}, and 8 hours for \textit{Opening} and \textit{Inserting}. 

During training, we normalized the channels for visual observation by: normalize visual representation to $(-1, 1)$, clip $(x, y, z)$ to the min and max workspace bounds, and set depth for all out-of-bound points to 0. 
Additionally, we append channels according to pixel location, following the original implementations in RVT. We apply random cropping augmentation to visual input during each training step.

\subsection{Policy Learning in Real-World Details (Sec \ref{sec:real})}

\textbf{Environment Setup.}
Our real-world evaluation platform uses a Franka arm mounted in a tabletop setup built with Vention frames.
Since the learned policy outputs 6-DoF end-effector poses and gripper actions, we use position control in all experiments, which is running at a fixed frequency of $20$ Hz. 
Specifically, given a target end-effector pose in the world frame, we first clip the pose to the pre-defined workspace. Then we perform linear interpolation from the current pose of the robot to the target pose with a step size of $5$mm for position and $1$ degree for rotation.
To move to each interpolated pose, we first calculate inverse kinematics to obtain the target joint positions based on current joint positions using the IK solver implemented in PyBullet~\cite{coumans2016pybullet}. Then we use the joint impedance controller from Deoxys~\cite{zhu2022viola} to reach to the target joint positions.
Two RGB-D cameras, Orbbec Femto Bolt, are mounted on the left side and the right side of the robot facing the workspace center. The cameras capture RGB images and point clouds at a fixed frequency of $20$ Hz.

\textbf{Tasks.}
We mirror the setup in simulation to evaluate on three similar tasks in the real-world: watering plant, opening drawer, and inserting pen into pen holder.
We collect a total of 10 demonstrations for each task and train a policy using the same training procedure described above.
The demonstrations are collected using kinesthetic teaching, consisting of varying numbers of keyframes (as described above) that are required to complete the task.
Success rates are visually examined by the operator.
10 trials with varying object configurations are performed, and the average success rate for each task is reported.

\subsection{Case Study: Scaling to Objaverse-XL Assets}

In this section, we describe an additional case study exploring the scalability of our pipeline to a more diverse set of 3D assets from Objaverse-XL~\cite{objaverseXL}.
\begin{compactitem}
    \item \textbf{Asset selection:} We selected objects from Objaverse-XL with LVIS category annotations, which represent common daily object categories. To ensure balanced representation across different categories, we randomly sampled 50 models from categories with abundant available models.
    \item \textbf{Asset filtering:} We filtered out assets with partially or fully transparent materials, as these produce incomplete depth renders required for accurate pointcloud reconstruction in our pipeline.
    \item \textbf{Asset rendering:} We rendered multi-view images of the assets using Blender with the official Objaverse codebase scripts. Assets where the rendering failed to fully capture the object were excluded from further processing.
    \item Following the asset preparation, we applied the same unsupervised affordance annotation extraction pipeline described in Section III-A to these Objaverse-XL assets.
\end{compactitem}

\end{document}